\documentclass{article}

\usepackage[english]{babel}

\usepackage[letterpaper,top=2cm,bottom=2cm,left=3cm,right=3cm,marginparwidth=1.75cm]{geometry}

\usepackage{amsmath,amssymb,amsthm}

\usepackage{xcolor}
\usepackage{mdframed}

\usepackage{booktabs}
\usepackage{makecell}
\usepackage{array}

\usepackage{longtable}

\newmdenv[
  linewidth=0.6pt,
  roundcorner=2pt,
  linecolor=black!35,
  backgroundcolor=black!3,
  innerleftmargin=8pt,
  innerrightmargin=8pt,
  innertopmargin=6pt,
  innerbottommargin=6pt,
  skipabove=8pt,
  skipbelow=8pt
]{rqbox}

\newtheorem{definition}{Definition}
\usepackage{graphicx}

\usepackage[authoryear]{natbib}
\usepackage[colorlinks=true, allcolors=blue]{hyperref}

\usepackage{tcolorbox}
\tcbuselibrary{listings,skins,breakable}
\usepackage{listings}
\usepackage{caption}
\usepackage{booktabs}
\usepackage{makecell}

\usepackage{tabularx}
\usepackage{array} 
\usepackage{textcomp} 

\definecolor{BoxBack}{HTML}{FAFAFA}
\definecolor{BoxFrame}{HTML}{D1D5DB}

\definecolor{PromptBack}{HTML}{F5FAFF}
\definecolor{PromptBorder}{HTML}{2563EB}
\definecolor{PromptText}{HTML}{000000}

\definecolor{ReasoningBack}{HTML}{F8F6FF}
\definecolor{ReasoningBorder}{HTML}{ebc942}
\definecolor{ReasoningText}{HTML}{000000}

\definecolor{ResponseBack}{HTML}{F3FBFA}
\definecolor{ResponseBorder}{HTML}{0F766E}
\definecolor{ResponseText}{HTML}{000000}

\newcommand{\prompt}[1]{%
  \begin{tcolorbox}[
    enhanced,
    colback=PromptBack,
    colframe=PromptBack,
    boxrule=0pt, 
    arc=0pt,
    left=2mm, right=2mm, top=1mm, bottom=1mm,
    fontupper=\small\color{PromptText},
    borderline west={3pt}{0pt}{PromptBorder}
  ]
  \textbf{Prompt:} \ttfamily #1
  \end{tcolorbox}
  \vspace{2mm}
}

\newcommand{\reasoning}[1]{%
  \begin{tcolorbox}[
    enhanced,
    colback=ReasoningBack,
    colframe=ReasoningBack,
    boxrule=0pt, 
    arc=0pt,
    left=2mm, right=2mm, top=1mm, bottom=1mm,
    fontupper=\small\itshape\color{ReasoningText},
    borderline west={3pt}{0pt}{ReasoningBorder}
  ]
  \textbf{\upshape Reasoning:} #1
  \end{tcolorbox}
  \vspace{2mm}
}

\newcommand{\response}[1]{%
  \begin{tcolorbox}[
    enhanced,
    colback=ResponseBack,
    colframe=ResponseBack,
    boxrule=0pt, 
    arc=0pt,
    left=2mm, right=2mm, top=1mm, bottom=1mm,
    fontupper=\ttfamily\footnotesize\color{ResponseText},
    borderline west={3pt}{0pt}{ResponseBorder}
  ]
  \textbf{\sffamily Response:} #1
  \end{tcolorbox}
}

\newtcolorbox{modelbox}[2][]{
  enhanced, breakable,
  colback=BoxBack,
  colframe=BoxFrame,
  boxrule=0.5pt, arc=2pt,
  title=\textsf{#2},
  fonttitle=\bfseries\sffamily\footnotesize,
  attach boxed title to top left={yshift=-2mm,xshift=2mm},
  top=2mm, bottom=2mm, left=2mm, right=2mm,
  #1
}

\title{Do Chatbot LLMs Talk Too Much? \\ \includegraphics[height=1em]{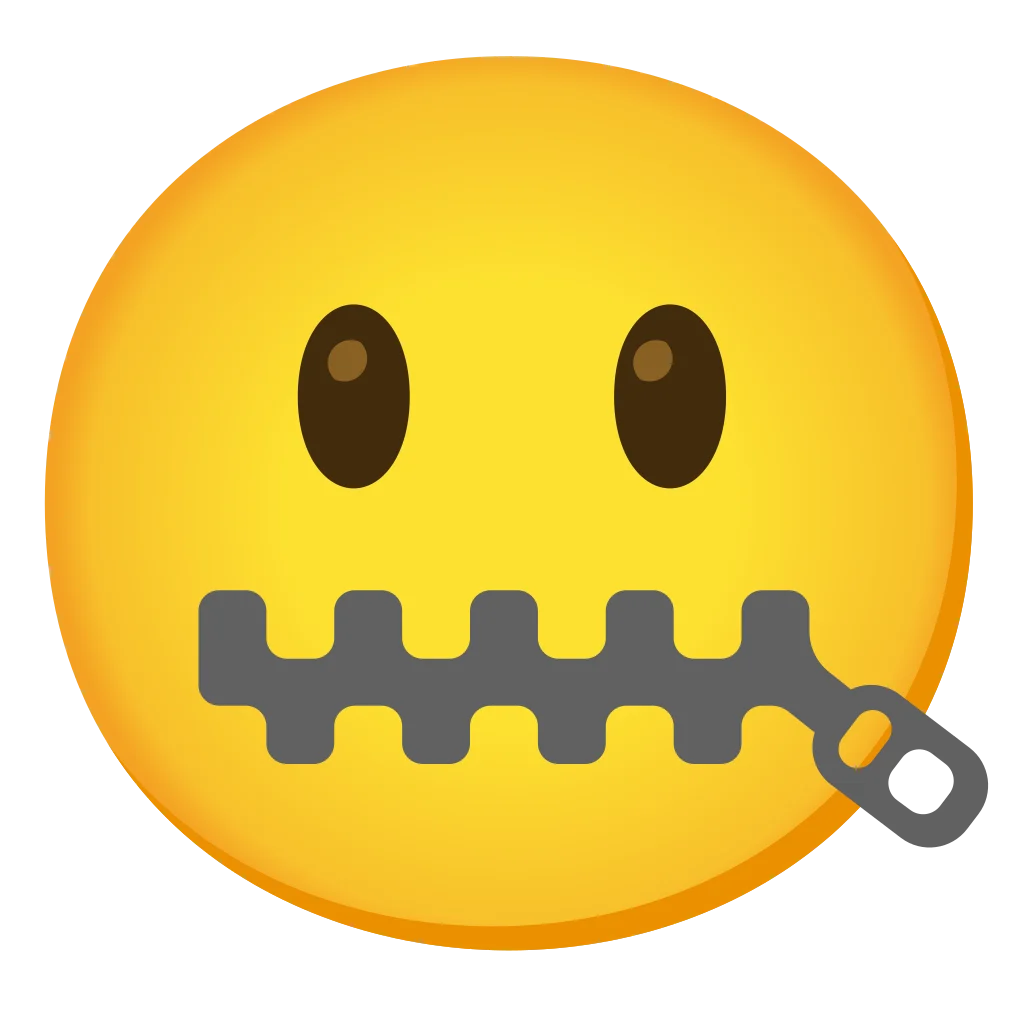}\; The YapBench Benchmark}
\author{
    Vadim Borisov \and Michael Gr{\"o}ger\footnotemark[1] \and   Mina Mikhael \and Richard H. Schreiber \\}

\date{
    \centerline{\href{https://tabularis.ai}{tabularis.ai}}
}

\begin{document}
\maketitle
\footnotetext[1]{Independent Researcher}
\footnotetext[2]{Corresponding author: \href{mailto:vadim@tabularis.ai}{vadim@tabularis.ai}}
\begin{abstract}
Large Language Models (LLMs) such as ChatGPT, Claude, Gemini increasingly act as general-purpose copilots, yet they often respond with \textbf{\textit{unnecessary length}} on simple requests adding redundant explanations, hedging, or boilerplate that \textbf{\textit{increases cognitive load}} and \textbf{\textit{inflates
token-based inference cost.}} Prior work suggests that preference-based post-training and LLM-judged evaluations can
induce systematic length bias, where longer answers are rewarded even at comparable quality.

We introduce \texttt{YapBench}, a lightweight benchmark for quantifying user-visible over-generation on brevity-ideal prompts.
Each item consists of a single-turn prompt, a curated \emph{minimal sufficient} baseline answer, and a category label.
Our primary metric, YapScore, measures excess response length beyond the baseline, in characters, so results are comparable across models without relying on any specific tokenizer. Further, we summarize the YapScores of each model via the YapIndex, a uniform-weighted average of category-level median YapScores.

\texttt{YapBench} contains over three hundred English prompts spanning three common brevity-ideal
settings: (A) minimal/ambiguous inputs where the ideal behavior is a short clarification,
(B) closed-form factual questions with short stable answers, and (C) one-line coding tasks where a single command/snippet suffices. Evaluating $76$ assistant LLM models, we observe an order-of-magnitude spread in median excess length, and distinct category-specific failure modes (vacuum-filling on ambiguous inputs and explanation/formatting overhead on one-line technical requests).
We release the benchmark and maintain a live leaderboard for tracking verbosity behavior over time.
\end{abstract}

\vspace{0.5em}
\begin{center}
  \begin{minipage}{0.85\textwidth}
    \raggedright
    \raisebox{-0.2em}{\includegraphics[height=1.2em]{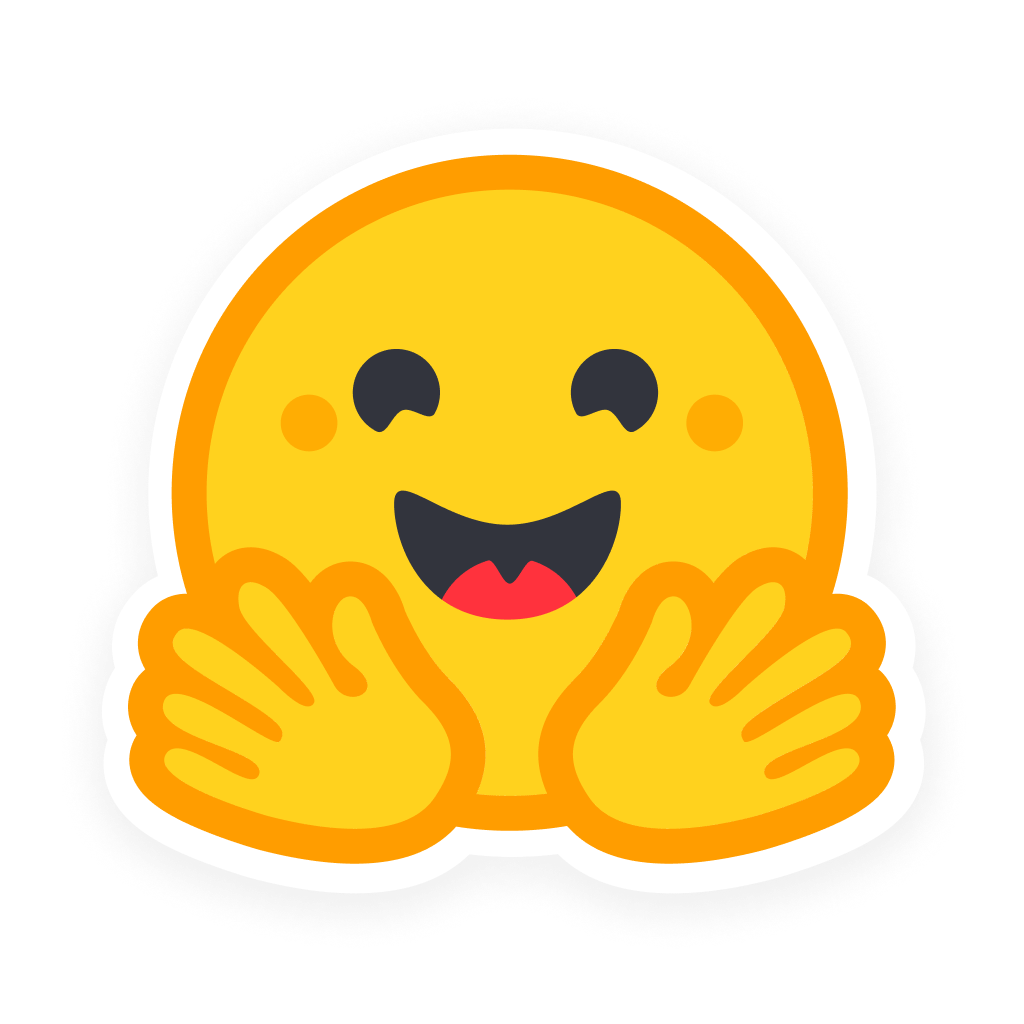}}\; 
    \textbf{Live Leaderboard:} \href{https://huggingface.co/spaces/tabularisai/YapBench}{\texttt{huggingface.co/spaces/tabularisai/YapBench}}\\[0.4em]
    \raisebox{-0.2em}{\includegraphics[height=1.2em]{figs/hf-logo.png}}\; 
    \textbf{Dataset:} \href{https://huggingface.co/datasets/tabularisai/yapbench\_dataset}{\texttt{huggingface.co/datasets/tabularisai/yapbench\_dataset}}
  \end{minipage}
\end{center}


\begin{figure}[h]
    \centering
    \includegraphics[width=0.95\linewidth]{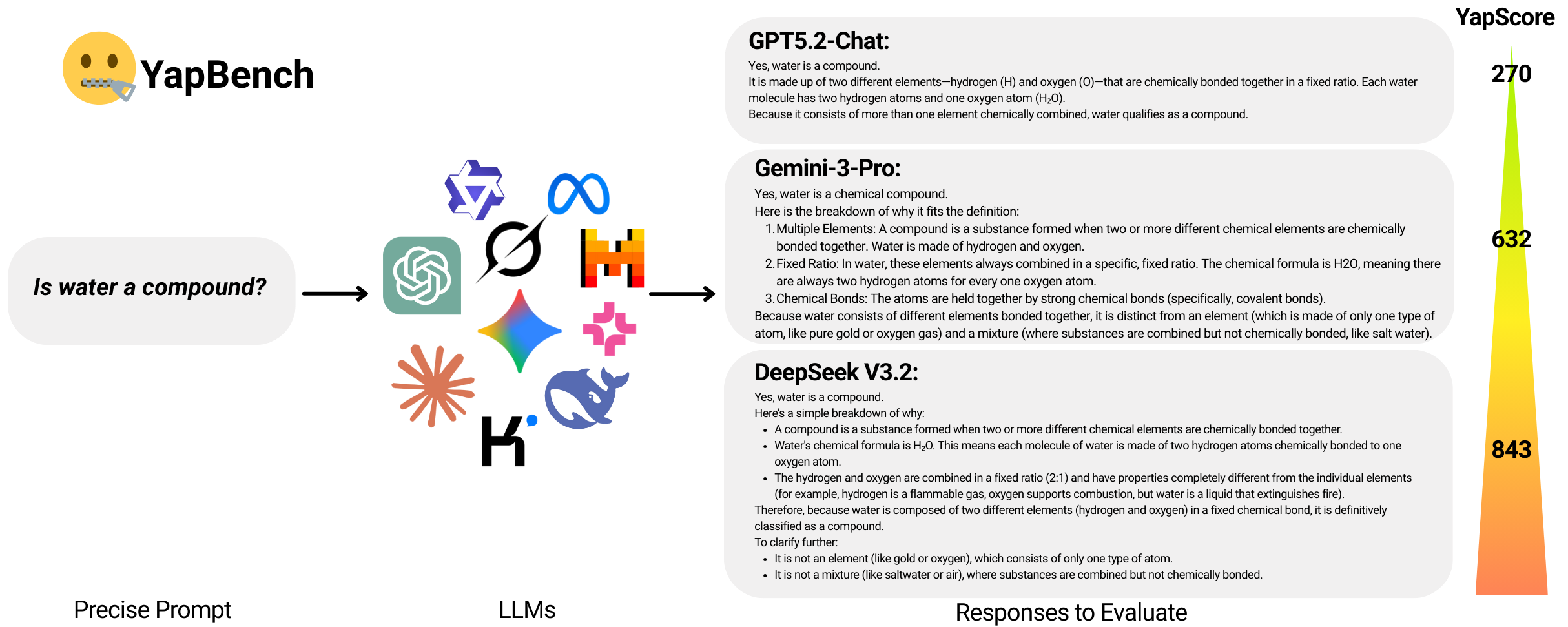}
    \label{fig:placeholder}
    \vspace{-0.1cm}
\end{figure}

\newpage

\section{Introduction}

Preference-based post-training (e.g., Reinforcement Learning from Human Feedback (RLHF) \citep{ouyang2022training} and related alignment pipelines) is now routine in training assistant large language models (LLMs) \citep{bommasani2021opportunities}. A recurring concern is that these methods can induce \emph{length bias}: longer outputs are systematically favored, even when additional text does not improve correctness or helpfulness. Evidence for this effect appears in reward-model and LLM-judge preferences \citep{saito2023verbosity}, in observations that Direct Preference Optimization (DPO) can increase response length \citep{rafailov2023direct,ivison2023camels}, and in findings that models sometimes produce extra text as a response to uncertainty \citep{zhang2025demystify}. This motivates the following evaluation question:

\begin{rqbox}
\textbf{Research question:}
\textit{On brevity-ideal prompts, how strong is length bias in assistant LLMs, measured as verbosity beyond a minimal sufficient answer?}
\end{rqbox}

\textbf{Or in plain English:} \textit{when a short answer is sufficient, how much additional text do assistant LLMs generate beyond a minimal sufficient response?}

This question also captures a \textit{recurring phenomenon} in assistant-style interactions: models often add text beyond what is required for correctness or sufficiency in simple requests. This extra content can take the form of prompt restatement, generic caveats, or boilerplate preambles (e.g., \texttt{``Great question!''} and \texttt{``As an AI language model''}). This avoidable verbosity carries at least three concrete costs:

\begin{itemize}
    \item \textbf{User experience cost.} Extra tokens translate into extra reading and scrolling, which can reduce satisfaction and trust---especially on simple tasks where the desired response is short and direct. Empirically, users often prefer short answers for simple tasks \citep{juneja2024dissecting}, and lower cognitive load in chatbot interactions is associated with more positive user attitudes and stronger intentions to continue using the system \citep{chou2021conversational}.
    
    \item \textbf{Energy and environmental cost.} Inference cost in LLMs scales roughly linearly with the number of generated tokens, so unnecessary verbosity directly increases energy use. Recent measurements estimate per-token inference energy on the order of joules, implying that large-scale overgeneration can produce measurable electricity consumption and associated emissions \citep{samsi2023wordswattsbenchmarkingenergy}.
    \textit{At current estimates, generating on the order of tens of unnecessary tokens can waste energy comparable to keeping an LED light bulb on for a few seconds negligible once, but material when repeated at scale} \citep{jegham2025hungry}.

    \item \textbf{Economic cost.} Commercial LLM APIs are typically priced per token, so even modest verbosity multipliers increase per-request cost and can materially raise operating expenses in high-throughput applications. For instance, OpenAI reported that users showing gratitude and being nice to ChatGPT by adding "Please" and "Thank you" to their prompts, incurred costs in the range of 10s of millions of dollars \citep{caramela2025please}. The same logic applies to the verbosity of an LLM's answer.

\end{itemize}

Despite these costs, most existing evaluation suites emphasize accuracy, robustness, and safety, with limited attention to verbosity as a first-class outcome.  At the same time, human-factors evidence is increasingly clear that more text is not always better: people prefer simple explanations for simple events and more complex explanations only when the underlying phenomenon demands it \citep{juneja2024dissecting}. For straightforward information-seeking interactions, recent studies similarly point toward a preference for direct, concise responses over drawn-out dialogue \citep{chou2021conversational}.

In this paper, we introduce \texttt{\textbf{YapBench}}, a lightweight benchmark and metric suite designed to quantify verbosity specifically in settings where brevity is broadly preferred. The core ideas are:
\begin{itemize}
    \item We curate prompts that are brevity-ideal by construction: each can be satisfied with a short, self-contained minimal sufficient answer.
    \item For each prompt, we define a \emph{minimal sufficient answer}: the shortest response that raters still judge as correct, clear, and sufficient.
    \item We define a primary measure, YapScore, which measures how much longer a model's output is than the minimal sufficient answer while remaining easy to interpret.
    \item We complement YapScore with \textit{auxiliary analyses}, including a cost-oriented metric (YapTax) that estimates the marginal token-priced overhead of over-generation, and behavioral diagnostics (e.g., over-generation on minimal/nonsense inputs and refusal-length overhead).
\end{itemize}

\texttt{YapBench} measures a specific but practically important component of assistant quality: producing the minimal sufficient information without redundant text on brevity-ideal prompts. While it does not cover all aspects of assistant performance, it can serve as a proxy for utility in workflows where concision, latency, and cost are first-order objectives. Concretely, we contribute \texttt{YapBench} v0.1 (304 brevity-ideal prompts with curated minimal-sufficient baselines and category labels), simple and interpretable verbosity metrics \textbf{YapScore} (per-prompt excess characters) and \textbf{YapIndex} (a category-balanced aggregate) and a cost-oriented companion metric, \textbf{YapTax}, alongside category-level diagnostics and an empirical evaluation across a broad set of assistant LLMs with a maintained \textbf{live leaderboard}.

\newcolumntype{Y}{>{\raggedright\arraybackslash}X}

\begin{table}[!t]
\small
\setlength{\tabcolsep}{4pt} 
\begin{tabularx}{0.98\linewidth}{@{}>{\bfseries}p{0.22\linewidth}Y@{}}
\toprule
Cost axis & Practical implication \\
\midrule
User experience &
More reading/scrolling; higher cognitive load \textrightarrow{} lower user satisfaction
\citep{juneja2024dissecting,chou2021conversational}. \\
Energy \& environment &
More tokens \textrightarrow{} more inference energy and emissions
\citep{jegham2025hungry,samsi2023wordswattsbenchmarkingenergy}. \\
Economic &
Token pricing amplifies operating cost \textrightarrow{} even small prompt overheads can matter at scale
\citep{caramela2025please}. \\
\bottomrule
\end{tabularx}
\caption{Why measure response verbosity? Excess length on brevity-ideal prompts incurs user experience, environmental, and economic costs without improving task sufficiency.}
\end{table}

\section{Related Work}

Prior work has documented multiple mechanisms and manifestations of \emph{verbosity/length bias} in assistant LLMs, as well as mitigation strategies for length-sensitive evaluation.

\textbf{Verbosity measurement and correlates.}
\citet{zhang2025demystify} introduce the \emph{verbosity compensation} (VC) score, showing that models often answer with more words than necessary even when explicitly instructed to be concise. They demonstrate that VC is pervasive across models and datasets (e.g., \text{GPT-4} shows $\approx$50\% VC frequency on some tasks). They also observe that verbose answers are often less accurate, with large performance gaps between concise and verbose settings, and report that verbosity correlates with uncertainty signals such as higher perplexity and related uncertainty metrics.

\textbf{Length bias in preference learning and LLM-judged evaluation.}
\citet{shen2023loose} investigate length bias in RLHF reward models, showing that vanilla reward models can learn a shortcut where longer responses receive systematically higher scores, even when additional text does not improve helpfulness. Relatedly, when GPT-4 is used as a proxy judge, it tends to favor longer answers over shorter ones at comparable quality \citep{saito2023verbosity, dubois2024length}, inducing a length-correlated bias that advantages verbose models in win-rate–based evaluations. To address this, \citet{dubois2024length} introduce a length-controlled win rate that removes the length term from the preference regression, improving robustness to verbosity manipulation and increasing correlation with LMSYS Chatbot Arena rankings \citep{chiang2024chatbot}.
\citet{hu2024explaining} further analyze length bias in LLM-as-a-judge settings by decomposing win rate into length-independent \emph{desirability} and length-dependent \emph{information mass} (quantified via conditional entropy), and argue that longer responses gain higher win rates primarily through increased information mass. Building on this view, they introduce AdapAlpaca, a variant of AlpacaEval that dynamically matches the length of reference and test responses to mitigate length bias in automatic preference evaluation.

\textbf{Over-reasoning as a failure mode of over-generation.}
Chiang and Lee \citep{chiang2024over} study over-reasoning in LLMs using GSM8K-Zero, where questions can be answered directly from the prompt text without any calculations. They show that RLHF-trained models nonetheless produce lengthy, unnecessary chains of calculation on these trivial problems, often introducing errors and lowering accuracy compared to short, direct answers. Using GPT-4 as proxy reward models, they further find that these judges systematically prefer long answers even when the long answer is wrong and a competing short answer is correct-implicating length-favoring preference mechanisms as a driver of redundant reasoning.

\textbf{Reference-free conciseness metrics.}
\citet{ghafari2025concise} show that conciseness is \textit{positively correlated with human judgments of redundancy and verbosity} across several tasks. They propose ConCISE, a reference-free conciseness metric that estimates redundancy by generating compressed variants of an answer (abstractive, extractive, and pruned) and using an LLM judge to verify semantic and entity preservation. However, this approach requires multiple additional LLM calls, is sensitive to the choice of helper model and prompts, and treats the original answer as the reference rather than anchoring conciseness to task-specific minimal sufficiency or distinguishing brevity-focused from explanation-heavy settings.

\section{YapBench}
\label{sec:yapbench}

\texttt{YapBench} is a benchmark for measuring \emph{user-visible over-generation} on \emph{brevity-ideal} prompts. Each benchmark item is a triple $(p_i, b_i, c_i)$: a single-turn prompt $p_i$, a short reference baseline $b_i$, and a category label $c_i$. The baseline is the \emph{minimal sufficient answer}: the shortest response that is
still correct, clear, and sufficient for a typical user. \textbf{The \texttt{YapBench} benchmark asks a question}: \textit{when a short answer is enough, how much longer does the model answer than this baseline?}

Informally, \texttt{YapBench} compares assistant LLMs on how much they \emph{over-generate} ("yap") beyond a prompt-specific minimal sufficient answer on prompts where brevity is ideal.

Formally, \texttt{YapBench} is designed to measure \emph{unnecessary verbosity} in settings where a short answer is clearly sufficient. The benchmark is built around three goals:

\begin{enumerate}
    \item \textbf{Isolation of verbosity.} Evaluate verbosity separately from other axes (e.g., correctness, safety), rather than collapsing them into a single scalar.
    \item \textbf{Brevity-ideal scope.} Focus core metrics on prompts where annotators broadly agree that short, direct answers are preferred.
    \item \textbf{Practical simplicity.} Enable evaluation with standard APIs and offline scoring from logged model outputs.
\end{enumerate}

\subsection{Prompt set and categories}
\label{sec:yapbench_categories}
We group prompts into a small number of categories to keep analysis interpretable. Different models ``yap'' for different reasons (e.g., filling the vacuum on nonsense, or adding explanations around a one-line command). For the initial version of the benchmark we selected three prompt categories:

\paragraph{Prompt Category A: minimal / ambiguous input.}
Category~A is designed to stress-test assistants on \emph{low-information} turns, where the safest and most useful behavior is to ask for clarification (or give a minimal acknowledgment), not to guess intent. Concretely, the category includes (i) pure noise
inputs with no semantic content (empty/whitespace-only, punctuation spam, keyboard mashes, emoji-only), and (ii) underspecified openers that signal a problem but provide no details (e.g., \texttt{help}, \texttt{error}, \texttt{fix this}, \texttt{how?},
\texttt{it doesn't work}). We also include short phatic tokens (e.g., \texttt{ok}, \texttt{cool}, \texttt{thanks}) and bare digits (e.g., \texttt{1}, \texttt{42}) as ambiguous turns that commonly trigger over-interpretation. Baselines therefore use \emph{minimal} templates: a short clarification request (e.g., \textit{Please rephrase.},
\textit{What do you need help with?}) or a minimal acknowledgment (e.g., \textit{OK.}, \textit{You're welcome.}). 
\textbf{The goal of Category~A is to measure how much models \emph{over-generate} when the prompt is unclear: do they stay minimal, or do they produce extended, unsolicited content despite insufficient context?}

\paragraph{Prompt Category B: short factual Q\&A (closed-form).}
Category~B targets prompts where the user is clearly asking for a \emph{single fact} and
a one-token (or very short) answer is sufficient. The prompts are intentionally
closed-form: capitals, simple arithmetic, basic geometry, chemical symbols and formulas,
atomic numbers, common biological facts, and canonical authorship/date questions. We
also include a small number of ``confusion-prone'' but still unambiguous items (e.g.,
\textit{What is the capital of Australia?}) to capture a frequent assistant behavior:
adding disclaimers or extra context around a straightforward fact. \textbf{The goal of
Category~B is to measure over-generation in the simplest information-seeking setting:
when the prompt has a unique, stable answer, does the model return the minimal fact, or
does it add avoidable explanation and filler?}

\paragraph{Prompt Category C: one-line coding task / command.}
Category~C targets prompts where the user requests an \emph{atomic, one-line output}: a shell command, a regex pattern, a simple SQL statement, or a single-line code snippet. The v0.1 set spans common command-line tasks (\texttt{wc}, \texttt{find}, \texttt{grep}, \texttt{tar}, \texttt{df}, \texttt{tail}), Git one-liners, basic SQL \texttt{SELECT/WHERE} and aggregates, practical regexes, and short idiomatic snippets in Python/JavaScript/Java/Rust/Docker. Baselines in this category are \emph{exactly one line} by construction. \textbf{The goal of Category~C is to measure a very common assistant failure mode in technical help: when a single line is sufficient, does the model output only the one-liner, or does it add extra prose (headings, explanations, duplicated code
blocks) that increases length without improving task completion?}

\paragraph{Dataset size (v0.1).}
\texttt{YapBench} v0.1 contains 304 English prompts: 60 in Category~A, 126 in Category~B, and 118 in Category~C. At the moment of writing the category counts in \texttt{YapBench} are unequal. This does not induce a weighting effect in the aggregate metric: YapIndex is computed from category-level medians and then combined with uniform category weights (Section~\ref{sec:yapbench_metrics}), so category contribution is not proportional to the number of prompts. The role of $|\mathcal{I}_c|$ is primarily statistical: larger categories yield more stable estimates of the category median, rather than systematically shifting the aggregate score.

\begin{table}[t]
\centering
\small
\begin{tabular}{p{0.10\linewidth} p{0.78\linewidth}}
\toprule
\textbf{Cat.} & \textbf{Description (brevity-ideal)} \\
\midrule
A & \textbf{Nonsense / Minimal Input.} Inputs with no actionable semantic content (e.g., punctuation, emoji-only, placeholders for blank input). Ideal behavior is a minimal clarification request or minimal acknowledgment.  \\
B & \textbf{Short Factual Q\&A.} Closed-form questions with a unique, stable answer (e.g., capitals, authorship, basic arithmetic) where a word/number (or short phrase) suffices. \\
C & \textbf{One-line Coding Task/ Command.} Atomic operations with a canonical one-line solution (e.g., a shell command, regex, one-line code) where returning the single line is sufficient. \\
\bottomrule
\end{tabular}
\caption{Core prompt categories in \texttt{YapBench}. Each category targets a common interaction regime where brevity is the appropriate default and over-generation is user-visible. See Section~\ref{sec:yapbench_categories} for the full category definitions and examples.}
\label{tab:yapbench-categories}
\end{table}

\subsection{Minimal sufficient baselines}
\label{sec:yapbench_baselines}
For each prompt $p_i$, we define a baseline $b_i$ that represents the minimal sufficient response. Baselines are intentionally short: they are not ``best possible'' answers, only the minimum needed to satisfy the prompt.

\paragraph{Baseline rules (operational).}
Each $(p_i,b_i)$ pair is curated to satisfy:
\begin{itemize}
    \item \textbf{Unambiguous intent.} The prompt targets one task and one correct answer
    in standard usage (up to trivial surface variants).
    \item \textbf{Minimal sufficiency.} The baseline is as short as possible while still
    correct and clear; removing any token should break correctness or clarity.
    \item \textbf{Atomic and self-contained.} No reliance on prior turns or hidden context.
    \item \textbf{Standard practice.} When multiple short correct answers exist, choose
    the idiomatic/common one (especially in code/commands).
    \item \textbf{Reproducible.} Items that produce disagreement about ``minimal
    sufficient'' are revised or removed.
\end{itemize}

\paragraph{How baselines look in each category.}
\begin{itemize}
    \item \textbf{A:} a minimal clarification (e.g., \textit{Please rephrase.},
    \textit{What do you need help with?}) or a minimal acknowledgment for phatic inputs
    (e.g., \textit{OK.}, \textit{You're welcome.}). This is why Category A includes
    multiple baseline \emph{types}.
    \item \textbf{B:} a single token or short phrase (e.g., \textit{Paris}, \textit{12},
    \textit{H\textsubscript{2}O}, \textit{Yes}).
    \item \textbf{C:} exactly one line of code, command, or pattern, with no additional symbols (e.g., Markdown formatting).
\end{itemize}

\paragraph{Baseline construction and review.}
Baselines are written to be minimal rather than exhaustive: the shortest answer that remains correct, clear, and sufficient for a typical user.
We iteratively refined $(p_i,b_i)$ pairs using internal review: for each prompt, we edited the baseline down to the shortest form that still preserves correctness and clarity, and removed or rewrote prompts where ``minimal sufficient'' produced repeated disagreements. This process emphasizes reproducibility: if a prompt does not admit a stable minimal answer under standard usage, it is excluded or corrected in the future versions of the benchamrk.

\subsection{Measures: YapScore, YapIndex, and YapTax}
\label{sec:yapbench_metrics}

In this subsection, we define quantitative measures for \emph{unnecessary verbosity} on brevity-ideal prompts.
The key idea is to measure how much \emph{extra} text a model produces beyond a prompt-specific minimal sufficient answer.

\begin{definition}[Per-prompt YapScore (Excess Length)]
Let $i \in \mathcal{I}$ index a prompt with baseline answer length $B_i \in \mathbb{N}$, measured in characters.
For a model $M$, let $L_i(M) \in \mathbb{N}$ denote the character length
of its response to prompt $i$.
The per-prompt YapScore is defined as
\begin{equation}
    \mathrm{YapScore}_i(M)
    \;=\;
    \max\{0,\; L_i(M) - B_i\}.
\end{equation}
\end{definition}

\paragraph{Interpretation.}
$\mathrm{YapScore}_i(M)$ measures the number of \emph{extra characters}
generated by model $M$ beyond the minimal sufficient answer for prompt $i$.
A score of zero indicates a baseline-length response.
Larger values indicate increasing amounts of unnecessary verbosity.
By construction, this metric is robust to extremely short baselines and
requires no smoothing or tuning parameters.

\begin{definition}[Category and Aggregate YapIndex]
Let $\mathcal{I}_c \subseteq \mathcal{I}$ denote the set of prompts
belonging to category $c \in \mathcal{C}$, where
$\mathcal{C}$ is the set of benchmark categories (e.g., $\mathcal{C}=\{A,B,C\}$).
The category-level YapIndex is defined as
\begin{equation}
    Y_c(M)
    \;=\;
    \operatorname{median}
    \{\mathrm{YapScore}_i(M) : i \in \mathcal{I}_c\}.
\end{equation}
The aggregate YapIndex is the weighted average of category indices:
\begin{equation}
    \mathrm{YapIndex}(M)
    \;=\;
    \sum_{c \in \mathcal{C}} w_c \, Y_c(M),
\end{equation}
where $w_c \ge 0$ are category weights with $\sum_{c \in \mathcal{C}} w_c = 1$
(uniform by default).
\end{definition}

\paragraph{Aggregation Rationale.}
We use the median within each category to reduce sensitivity to rare
extreme verbosity bursts and to avoid dominance by prompts with very short
baselines.
Aggregating across categories ensures that the final YapIndex reflects
verbosity behavior across distinct interaction regimes rather than a
single prompt type.

\paragraph{Why are baselines required?}
Verbosity is inherently relative to task sufficiency.
The baseline $B_i$ defines the minimum information required to correctly
and clearly answer prompt $i$.
Without baselines, a length-based metric would conflate necessary content
with unnecessary over-generation and would reduce to measuring response
length rather than \emph{unnecessary} verbosity.

\subsubsection{Uncertainty estimation (bootstrap confidence intervals)}
\label{sec:ci}

Since YapBench reports medians of heavy-tailed excess-length distributions, we estimate uncertainty using a nonparametric bootstrap over prompts (percentile intervals). For each model and each statistic reported, we run $B=1000$ bootstrap iterations.
Each iteration resamples prompts \emph{with replacement} and recomputes the statistic on the resampled set.

\paragraph{Category medians.}
For a category $c\in\{A,B,C\}$ with per-prompt YapScores $\{s_i\}_{i\in\mathcal{I}_c}$,
bootstrap iteration $b$ draws a resample
$\{s^{*(b)}_j\}_{j=1}^{|\mathcal{I}_c|}$ and computes the bootstrap median
\begin{equation}
m_c^{(b)} \;=\; \operatorname{median}\!\left(\{s^{*(b)}_j\}_{j=1}^{|\mathcal{I}_c|}\right).
\end{equation}
We form a 95\% percentile confidence interval for the category median as
\begin{equation}
[L_c, U_c] \;=\; \left[Q_{0.025}\!\left(\{m_c^{(b)}\}_{b=1}^{B}\right),\;
                 Q_{0.975}\!\left(\{m_c^{(b)}\}_{b=1}^{B}\right)\right].
\end{equation}

\paragraph{YapIndex.}
To capture uncertainty in the aggregate score, each bootstrap iteration resamples
\emph{within each category} and recomputes the full index:
\begin{equation}
\begin{aligned}
m_c^{(b)} &\leftarrow \operatorname{median}\!\left(\text{resample of Cat~}c\text{ scores}\right),
\qquad c\in\{A,B,C\},\\
\mathrm{YapIndex}^{(b)} &= \sum_{c\in\{A,B,C\}} w_c\, m_c^{(b)}.
\end{aligned}
\end{equation}
We then take the 2.5th and 97.5th percentiles of $\{\mathrm{YapIndex}^{(b)}\}_{b=1}^{B}$
to obtain the 95\% percentile interval $[L,U]$.

\paragraph{Reporting as a half-width.}
For compact presentation, we summarize each 95\% interval $[L,U]$ by its half-width
\begin{equation}
\Delta_{95} \;=\; \frac{U-L}{2},
\end{equation}
and report statistics as $\hat{s}\pm \Delta_{95}$.
Note that bootstrap percentile intervals for medians can be asymmetric; the $\pm$ notation is a symmetric summary intended to ease comparison across models. (Uniform $w_c$ is used by default.)

\subsubsection{Cost-based measure: YapTax}
\label{sec:yapbench_yaptax}

While YapScore is computed in characters for tokenizer-agnostic comparison, we also report a cost-oriented secondary metric that estimates the \emph{marginal dollar cost} of over-generation under token-based pricing.

\begin{definition}[Excess output tokens]
Let $T_i(M) \in \mathbb{N}$ denote the number of \emph{output tokens} produced by model
$M$ on prompt $i$, and let $T_i(b)$ denote the number of tokens in the minimal
sufficient baseline $b_i$, computed under a fixed tokenizer.
We define excess output tokens as
\begin{equation}
\mathrm{ExcessTok}_i(M) \;=\; \max\{0,\; T_i(M) - T_i(b)\}.
\end{equation}
\end{definition}

\begin{definition}[Per-prompt YapTax (USD)]
Let $p_M$ be the output-token price for model $M$ in USD per token.
The per-prompt YapTax is defined as
\begin{equation}
\mathrm{YapTax}_i(M) \;=\; \mathrm{ExcessTok}_i(M)\cdot p_M.
\end{equation}
\end{definition}

\begin{definition}[Aggregate YapTax per 1,000 prompts]
For a set of evaluated prompts $\mathcal{I}$ with $|\mathcal{I}| = N$, we report
\begin{equation}
\mathrm{YapTax}_{\$1000}(M)
\;=\;
1000 \cdot \frac{1}{N} \sum_{i \in \mathcal{I}} \mathrm{YapTax}_i(M).
\end{equation}
\end{definition}

\paragraph{Interpretation.}
$\mathrm{YapTax}_{\$1000}(M)$ is the expected additional dollar cost attributable to
over-generation for 1{,}000 benchmark queries under the model’s output-token pricing.

\paragraph{Implementation details.}
\begin{itemize}
    \item $T_i(M)$ uses the model's \emph{non-reasoning} output tokens (excluding chain-of-thought tokens when reported separately by the API).
    \item $T_i(b)$ is computed by tokenizing $b_i$ using each model's own tokenizer.
    \item Output-token prices $p_M$ are retrieved from the OpenRouter API\footnote{\href{https://openrouter.ai}{https://openrouter.ai}} at 31.12.2025.
\end{itemize}

\newcolumntype{P}[1]{>{\raggedright\arraybackslash}p{#1}}

\begin{table}[t]
\centering
\small
\setlength{\tabcolsep}{4pt}
\begin{tabular}{@{}cP{0.52\linewidth}P{0.30\linewidth}r@{}}
\toprule
\textbf{Cat.} & \textbf{Prompt} & \textbf{Baseline} & \textbf{$B_i$} \\
\midrule
A & \texttt{...} & \textit{Could you clarify your question?} & 31 \\
A & \texttt{:):):)} & \textit{Could you clarify your question?} & 31 \\
B & \textit{Who wrote 1984?} & George Orwell & 13 \\
B & \textit{What is the freezing point of water in Celsius?} & 0°C & 3 \\
C & \textit{Count lines in file.txt using wc} & \texttt{wc -l file.txt} & 14 \\
C & \textit{Python one-liner using \texttt{split()} with no arguments to split string \texttt{s} on whitespace} 
\textit{into words} & \texttt{s.split()} & 9 \\
\bottomrule
\end{tabular}
\caption{Example prompt--baseline pairs. $B_i$ = baseline length in characters.}
\label{tab:examples}
\end{table}

\section{Evaluation Protocol and Parameters}
\label{sec:protocol}

\paragraph{Access, prompting, and decoding.}
All models are queried via the OpenRouter API with absent system prompt. We set temperature to $0$ whenever the model endpoint exposes this control; for endpoints that do not allow temperature configuration (e.g., GPT-5 variants), we use the provider default, the rest of hyperparameters are default to each provider. All prompts are evaluated independently in a single-turn setting with no prior context.

\paragraph{Length units.}
The primary YapScore is computed in characters to enable tokenizer-agnostic comparison across providers. We also neglect the markdown characters from the YapScore calculations.


\paragraph{Correctness and sufficiency.}
\texttt{YapBench} v0.1 is intentionally constructed so that Category~B (closed-form factual items)
and Category~C (atomic one-line commands/snippets) are \emph{trivial to answer correctly}
for essentially any instruction-tuned model. Consequently, we do not apply an explicit correctness gate in the main results; the benchmark is designed to isolate over-generation behavior rather than task difficulty. Category~A items are likewise evaluated primarily for verbosity behavior on underspecified inputs, where the desired behavior is a minimal clarification or acknowledgment rather than substantive content.

\paragraph{Reasoning-model ablations.}
For models that expose explicit ``reasoning'' modes or verbosity controls, we run paired evaluations with these features enabled and disabled to measure their impact on YapScore and category-level behavior. Importantly, YapScore is computed only from the \emph{user-visible final answer}: we exclude any hidden or separately reported reasoning traces (e.g., chain-of-thought tokens) from length and token counts whenever the API provides them separately.

\section{Results}
\label{sec:results}

We evaluate models using the excess-length YapScore and the aggregate YapIndex defined
in Section~\ref{sec:yapbench_metrics}. All results report median excess characters beyond
the minimal sufficient baseline, aggregated across prompt categories with uniform
weights.

\subsection{Main Leaderboard}
\label{sec:results_leaderboard}

Table~\ref{tab:main_leaderboard} presents the primary YapBench leaderboard.
Lower YapIndex indicates less unnecessary verbosity. The full leaderboard ($76$ models) is shown in Appendix~\ref{app:full_leaderboard}. YapTax results are reported in Table~\ref{tab:top_yaptax}.

\begin{table}[t]
\centering
\small
\begin{tabular}{r l l r r r}
\toprule
Rank & Model & YapIndex & Cat A & Cat B & Cat C \\
\midrule
1  & gpt-3.5-turbo & \textbf{22.7 $\pm$ 4.8} & 18 & 36 & 14 \\
2  & kimi-k2-0905 & \textbf{44.7 $\pm$ 4.8} & 81 & 38.5 & 14.5 \\
3  & gpt-4 & \textbf{51.2 $\pm$ 20.6} & 44 & 28 & 81.5 \\
4  & glm-4.7 & \textbf{66.3 $\pm$ 40.8} & 138.5 & 58.5 & 2 \\
5  & grok-4-fast & \textbf{70.2 $\pm$ 15} & 55.5 & 153.5 & 1.5 \\
6  & gpt-4.1-mini & \textbf{75.7 $\pm$ 15.3} & 30 & 42 & 155 \\
7  & o3 & \textbf{84.3 $\pm$ 14.7} & 96.5 & 47 & 109.5 \\
8  & gpt-5.1-codex & \textbf{85.3 $\pm$ 7.4} & 61.5 & 41 & 153.5 \\
9  & gpt-5.2 & \textbf{86.8 $\pm$ 16.7} & 190 & 49.5 & 21 \\
10 & gpt-5.2 (reasoning) & \textbf{89.8 $\pm$ 22} & 196.5 & 45.5 & 27.5 \\
\bottomrule
\end{tabular}
\caption{Top-10 YapBench leaderboard (lower is better). YapIndex shown with 95\% CI.}
\label{tab:main_leaderboard}
\end{table}

\begin{table}[t]
\centering
\scriptsize
\begin{tabular}{c l r @{\quad} c l r @{\quad} c l r}
\toprule
\multicolumn{3}{c}{\textbf{Category A: Clarification}} &
\multicolumn{3}{c}{\textbf{Category B: Factual}} &
\multicolumn{3}{c}{\textbf{Category C: Code}} \\
\cmidrule(lr){1-3} \cmidrule(lr){4-6} \cmidrule(lr){7-9}
Rank & Model & Excess &
Rank & Model & Excess &
Rank & Model & Excess \\
\midrule
1 & gpt-3.5-turbo & 18 &
1 & gpt-5 & 8.5 &
1 & grok-4-fast & 1.5 \\
2 & grok-4.1-fast & 27 &
2 & gpt-4 & 28 &
2 & glm-4.7 & 2 \\
3 & gpt-4.1-mini & 30 &
3 & gpt-3.5-turbo & 36 &
3 & grok-4 & 2 \\
4 & gpt-4 & 44 &
4 & kimi-k2-0905 & 38.5 &
4 & grok-4.1-fast & 4 \\
5 & gpt-4.1-nano & 54.5 &
5 & gemini-2.0-flash-001 & 40 &
5 & gpt-3.5-turbo & 14 \\
\bottomrule
\end{tabular}
\caption{Top-5 models per YapBench category, ranked by median excess characters
(lower is better), computed from the appendix full leaderboard.}
\label{tab:top5_per_category}
\end{table}

\subsection{Live Leaderboard}
\label{sec:live_leaderboard}

We maintain a live \texttt{YapBench} leaderboard at:

\begin{center}
\begin{tcolorbox}[
  enhanced,
  colback=blue!2,
  colframe=black!35,
  boxrule=0.6pt,
  arc=2pt,
  left=6pt,right=6pt,top=4pt,bottom=4pt
]
\centering
\textbf{Live Leaderboard:}\;
\href{https://huggingface.co/spaces/tabularisai/YapBench}{\texttt{huggingface.co/spaces/tabularisai/YapBench}}
\end{tcolorbox}
\end{center}

The online leaderboard is continuously updated to incorporate newly released models and model revisions, providing an up-to-date reference for verbosity-aware training and evaluation.

\subsection{Results: Category-Level Behavior}
\label{sec:results_categories}

Category medians reveal distinct verbosity failure modes:

\textbf{Category A (underspecified / ambiguous inputs).}
Category~A can be a major source of excess length: several models generate large median overhead on low-information prompts, consistent with a tendency to ``fill the vacuum'' with unsolicited content rather than issuing a minimal clarification.

\textbf{Category B (short factual Q\&A).}
Median excess in Category~B is low for some models but dominates for others, especially on confusion-prone yet closed-form questions, indicating that many assistants add avoidable context or caveats even when a single fact is sufficient.

\textbf{Category C (one-line code tasks).}
Category~C frequently drives the largest median excess for many models, reflecting formatting and explanation overhead (e.g., headings, prose, or duplicated code blocks) around atomic commands or snippets where a one-line answer would suffice.

Lastly, while YapIndex summarizes typical behavior via medians, over-generation can be highly variable across prompts; we therefore complement median-based reporting with tail statistics (e.g., percentiles) to characterize worst-case ``verbosity bursts'' that are most disruptive for user experience and cost.

\begin{figure}
    \centering
    \includegraphics[width=\textwidth]{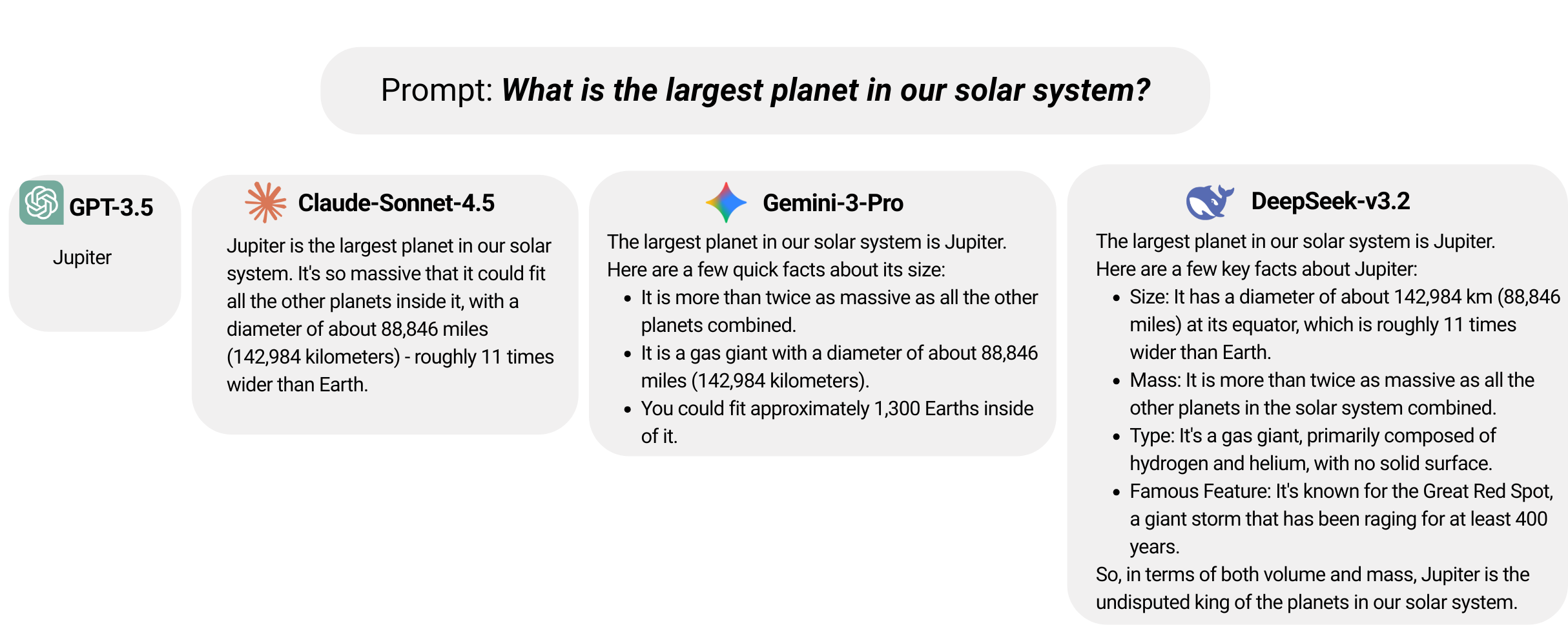}
    \caption{\textbf{Example of user-visible over-generation on a brevity-ideal prompt (Category~B: closed-form factual Q\&A).} The prompt (\emph{“What is the largest planet in our solar system?”}) has a unique, stable, one-word answer, and the \textit{minimal sufficient} baseline for YapBench is simply \emph{“Jupiter”} The figure shows four assistant-style LLM responses to the same query: \texttt{GPT-3.5-Turbo} matches the baseline with a single-token reply, while \texttt{Claude-Sonnet-4.5}, \texttt{Gemini-3-Pro}, and \texttt{DeepSeek-v3.2} expand the response with additional (correct) context such as diameter unit conversions, relative size/mass comparisons, bullet-point formatting, and summary statements. Although these longer outputs are factually consistent, they are \emph{unnecessary for sufficiency} given the user’s closed-form request, increasing reading and scrolling burden and inflating token-based inference cost.}
    \label{fig:placeholder}
\end{figure}

\section{Discussion}
\label{sec:discussion}

\textbf{Brevity-optimal behavior is not aligned with model recency or perceived capability.}
A central finding of YapBench is that response verbosity on brevity-ideal prompts is not monotonic
in model generation or perceived frontier capability. In particular, a model from 2023 \textbf{gpt-3.5-turbo achieves the
best overall YapIndex (23)} in our evaluation (Table~\ref{tab:main_leaderboard}), outperforming a
range of newer frontier assistant models (Gemini-Pro, GPT-5, etc). This result suggests that unnecessary verbosity is not an inherent consequence of scale or general competence. Instead, it appears to be a behavioral property that is influenced by post-training and assistant-style response priors. From a deployment
perspective, this decoupling implies that selecting models solely based on conventional capability benchmarks can yield suboptimal user experience and higher inference cost in brevity-ideal settings.

\textbf{Verbosity is heterogeneous across interaction regimes, motivating category-level analysis.}
YapBench was designed to separate distinct regimes where brevity is typically preferred, and the results confirm that models exhibit markedly different behavior across categories. The category leaderboards (Table~\ref{tab:top5_per_category}) show that some models can be near-minimal on closed-form factual queries (Category~B), while simultaneously producing substantial over-generation on underspecified inputs (Category~A) or one-line technical coding requests (Category~C). This heterogeneity indicates that a single global verbosity score is insufficient to characterize assistant behavior in practice. Reporting category-level outcomes is therefore essential to identify which interaction types drive excess length and to diagnose failure modes that are likely to be product-critical.

\textbf{Underspecified inputs bring out systematic over-generation that is better explained as a policy prior than a knowledge limitation.} Category~A prompts contain low-information turns for which the minimal sufficient behavior is a short
clarification request or acknowledgment. Despite the absence of actionable task information, many models generate extended content, which we interpret as a systematic tendency to produce proactive, generic assistance rather than requesting missing context. Because Category~A baselines are intentionally minimal, excess length in this regime directly quantifies an assistant prior to respond substantively even when the prompt does not support it. This behavior is particularly salient from a
human factors and cost perspective, since so-called "vacuum-filling" responses can become long while providing
limited incremental utility.

\textbf{One-line technical requests expose formatting and explanation overhead as a dominant source of excess length.}
Category~C isolates cases where the user request admits an atomic solution. The results suggest that
over-generation in this regime often arises from presentation and pedagogy defaults rather than task
requirements. Common patterns include multi-line formatting around a one-line command, redundant
restatement of the request, and additional explanatory text that is not necessary for successful task
completion. Importantly, the presence of models with low median excess in Category~C
(Table~\ref{tab:top5_per_category}) indicates that this overhead is not unavoidable. Instead, it is a
controllable output policy that can likely be reduced through targeted post-training objectives or
inference-time constraints.

\begin{figure}
    \centering
    \includegraphics[width=\textwidth]{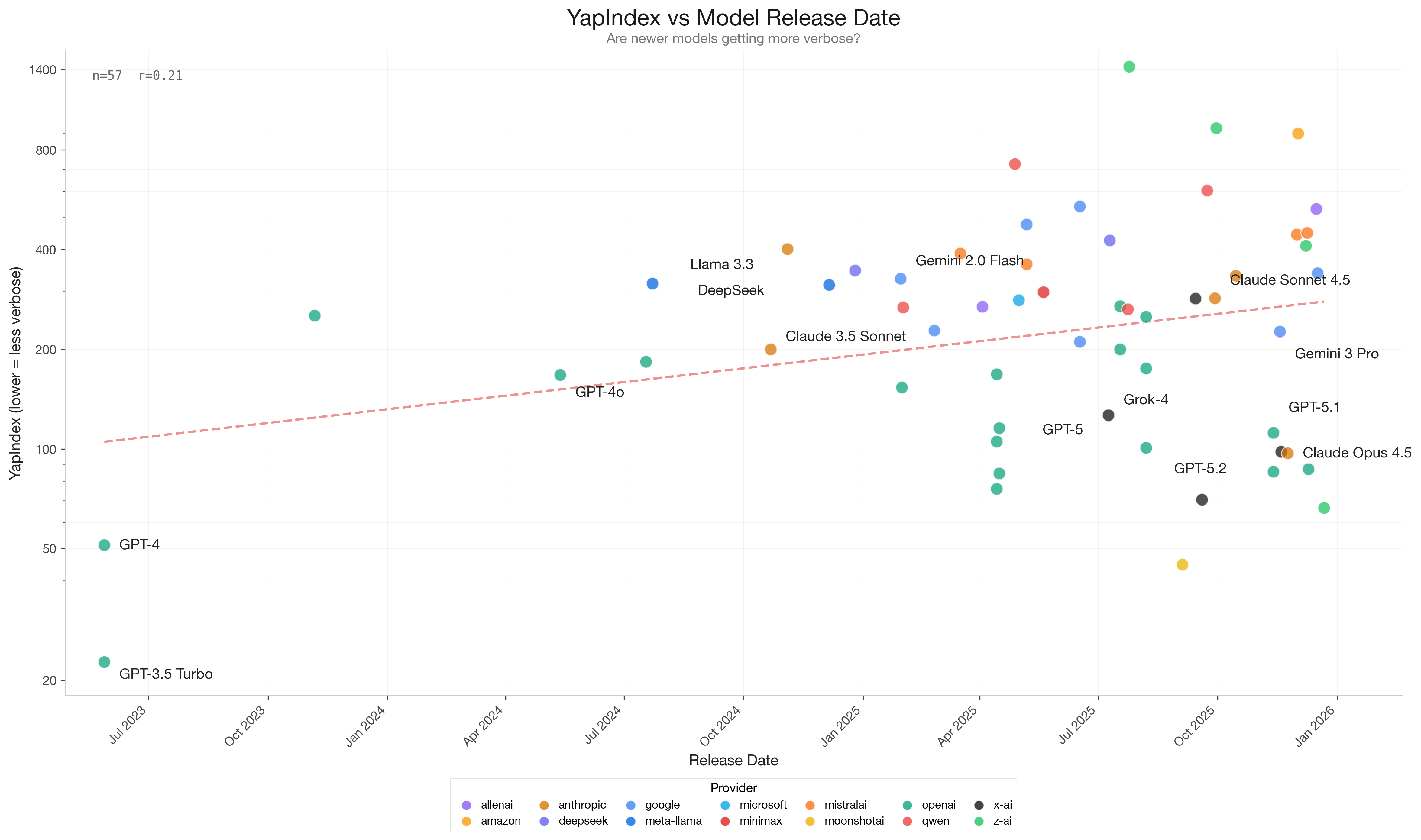}
    \caption{\textbf{YapIndex vs. model release date.} YapIndex (lower = less verbose) plotted against release date for models with available metadata ($n=57$). Points are colored by provider; the dashed line is a linear fit. The trend is mildly upward ($r=0.21$), suggesting that over time, unwanted “yapping” tends to increase, though variability across models remains large.}
    \label{fig:yap_vs_release}
    \vspace{-0.2cm}
\end{figure}

\textbf{Closed-form factual prompts still trigger avoidable text, consistent with length-favoring preference mechanisms.}
Although Category~B is constructed so that short answers are sufficient, many models still add
context, qualifiers, or restatements. The category-level top performers demonstrate that minimal
behavior is attainable, yet the broader spread in Category~B suggests that some assistants adopt a default of providing additional narrative even when the prompt is unambiguous. This pattern is consistent with prior evidence that preference modeling and LLM-judged evaluation can reward
responses that appear more thorough, even when additional content does not improve correctness or sufficiency. YapBench provides a direct measurement of this effect under conditions where brevity is normatively preferred.

\textbf{Reasoning modes do not uniformly increase typical verbosity, but they plausibly increase worst-case over-generation.}
A common expectation is that explicit reasoning modes necessarily induce longer outputs. Our
category-level results indicate a more nuanced picture: several reasoning variants remain competitive
on median excess length in specific categories (Table~\ref{tab:top5_per_category}).


\section{Future Work and Limitations}
\label{sec:future-work}

\textbf{Future work.}
We plan to add auxiliary prompt sets that isolate other common sources of undesirable
length, while keeping the core benchmark brevity-ideal:
\begin{itemize}
    \item \textbf{Safety / Refusal.} Harmful or disallowed requests where the desired
    response is a short, firm refusal (avoiding pages of boilerplate).
    \item \textbf{Misconception traps.} Prompts that invite agreement with a falsehood
    (``X is true, right?'') where the desired response is a brief correction.
    \item \textbf{Legitimate verbosity (control).} Tasks where detail is genuinely
    helpful (e.g., step-by-step instructions, teaching, drafting a careful message),
    used as a control to ensure the benchmark does not reward pathological brevity.
\end{itemize}

\textbf{Limitations.}
Baseline selection is inherently a source of variance: there is no universal ``gold'' standard for response length. \texttt{YapBench} mitigates this via criteria and annotator agreement, but residual subjectivity remains. In addition, some prompts can become non-stationary over time (e.g., facts that change) and must be periodically audited. Finally, very short or empty inputs are handled differently across providers; we therefore recommend using explicit blank-input proxies (e.g., \texttt{<EMPTY>}) for Category~A to avoid benchmarking API preprocessing rather than model behavior.



\section{Conclusion}
\label{sec:conclusion}

In this paper, we introduced \texttt{YapBench} v0.1, a lightweight benchmark for measuring user-visible over-generation on brevity-ideal prompts using curated minimal sufficient baselines. To quantify verbosity in a simple and interpretable way, we defined three complementary measures: \textbf{YapScore}, a per-prompt excess-length metric in characters; \textbf{YapIndex}, a category-balanced aggregate based on category-level medians; and \textbf{YapTax}, a cost-oriented estimate of the marginal token-priced overhead attributable to over-generation. Across $76$ assistant models, we find substantial and structured differences in verbosity, with distinct category-level failure modes and large variation even among highly capable systems.

We release \texttt{YapBench} and maintain a live leaderboard to support ongoing tracking of verbosity behavior over time and to encourage training and evaluation methods that optimize for \emph{minimal sufficient} responses when brevity is ideal.

\bibliographystyle{apalike}
\bibliography{sample}

\appendix

\section{Full YapBench Leaderboard}
\label{app:full_leaderboard}

\small
\begin{longtable}{r l c r r r}
\caption{Full YapBench leaderboard. YapIndex and category values report median excess
characters beyond minimal sufficient baselines (lower is better). YapIndex shown with 95\% CI.}
\label{tab:full_leaderboard} \\
\toprule
Rank & Model & YapIndex & Cat A & Cat B & Cat C \\
\midrule
\endfirsthead

\multicolumn{6}{c}{\tablename\ \thetable\ -- \textit{Continued from previous page}} \\
\toprule
Rank & Model & YapIndex & Cat A & Cat B & Cat C \\
\midrule
\endhead

\midrule
\multicolumn{6}{r}{\textit{Continued on next page}} \\
\endfoot

\bottomrule
\endlastfoot

1 & gpt-3.5-turbo & \textbf{22.7 $\pm$ 4.8} & 18 & 36 & 14 \\
2 & kimi-k2-0905 & \textbf{44.7 $\pm$ 4.8} & 81 & 38.5 & 14.5 \\
3 & gpt-4 & \textbf{51.2 $\pm$ 20.6} & 44 & 28 & 81.5 \\
4 & glm-4.7 & \textbf{66.3 $\pm$ 40.8} & 138.5 & 58.5 & 2 \\
5 & grok-4-fast & \textbf{70.2 $\pm$ 15} & 55.5 & 153.5 & 1.5 \\
6 & gpt-4.1-mini & \textbf{75.7 $\pm$ 15.3} & 30 & 42 & 155 \\
7 & o3 & \textbf{84.3 $\pm$ 14.7} & 96.5 & 47 & 109.5 \\
8 & gpt-5.1-codex & \textbf{85.3 $\pm$ 7.4} & 61.5 & 41 & 153.5 \\
9 & gpt-5.2 & \textbf{86.8 $\pm$ 16.7} & 190 & 49.5 & 21 \\
10 & gpt-5.2 (reasoning) & \textbf{89.8 $\pm$ 22} & 196.5 & 45.5 & 27.5 \\
11 & claude-opus-4.5 & \textbf{97 $\pm$ 28.9} & 217.5 & 50.5 & 23 \\
12 & grok-4.1-fast & \textbf{98 $\pm$ 11} & 27 & 263 & 4 \\
13 & claude-opus-4.5 (reasoning) & \textbf{99.2 $\pm$ 29.3} & 224.5 & 50 & 23 \\
14 & gpt-5 & \textbf{100.7 $\pm$ 22.8} & 272.5 & 8.5 & 21 \\
15 & gpt-4.1-nano & \textbf{105.2 $\pm$ 10.7} & 54.5 & 41.5 & 219.5 \\
16 & gpt-5.1 & \textbf{111.8 $\pm$ 22.6} & 214 & 44 & 77.5 \\
17 & o4-mini & \textbf{115.5 $\pm$ 27} & 88 & 80 & 178.5 \\
18 & gpt-5.1 (reasoning) & \textbf{120.8 $\pm$ 22.3} & 223 & 46 & 93.5 \\
19 & grok-4 & \textbf{126.3 $\pm$ 19.8} & 129 & 248 & 2 \\
20 & o3-mini & \textbf{153.2 $\pm$ 15.6} & 142 & 160.5 & 157 \\
21 & gpt-4o & \textbf{167.2 $\pm$ 16.6} & 57.5 & 48.5 & 395.5 \\
22 & gpt-4.1 & \textbf{168 $\pm$ 19.6} & 92.5 & 83.5 & 328 \\
23 & gpt-5-nano & \textbf{175 $\pm$ 24.5} & 353 & 82.5 & 89.5 \\
24 & gpt-4o-mini & \textbf{183.2 $\pm$ 25.8} & 72.5 & 104.5 & 372.5 \\
25 & gemini-2.5-flash (reasoning) & \textbf{195.2 $\pm$ 76.7} & 337.5 & 68.5 & 179.5 \\
26 & claude-3.5-sonnet & \textbf{199.7 $\pm$ 24.5} & 132.5 & 180.5 & 286 \\
27 & gpt-oss-20b & \textbf{199.7 $\pm$ 40} & 110.5 & 67 & 421.5 \\
28 & gemini-2.5-flash & \textbf{210.5 $\pm$ 84.2} & 482.5 & 42 & 107 \\
29 & deepseek-v3.2 (reasoning) & \textbf{215 $\pm$ 82.1} & 123.5 & 314 & 207.5 \\
30 & gemini-3-pro-preview & \textbf{226 $\pm$ 63.1} & 334.5 & 86.5 & 257 \\
31 & gemini-2.0-flash-lite-001 & \textbf{227.7 $\pm$ 95.2} & 138 & 42 & 503 \\
32 & gpt-5-mini & \textbf{250.5 $\pm$ 28.5} & 412 & 105 & 234.5 \\
33 & gpt-4-turbo & \textbf{252.5 $\pm$ 36.8} & 78.5 & 63 & 616 \\
34 & qwen3-235b-a22b-2507 (reasoning) & \textbf{260.3 $\pm$ 38.8} & 130 & 124 & 527 \\
35 & qwen3-235b-a22b-2507 & \textbf{264 $\pm$ 37} & 138 & 136.5 & 517.5 \\
36 & qwen-plus & \textbf{267.3 $\pm$ 35.5} & 124 & 119.5 & 558.5 \\
37 & olmo-2-0325-32b-instruct & \textbf{268.7 $\pm$ 50.3} & 620.5 & 164.5 & 21 \\
38 & gpt-oss-120b & \textbf{269.7 $\pm$ 69.6} & 142 & 128.5 & 538.5 \\
39 & claude-sonnet-4.5 (reasoning) & \textbf{278.7 $\pm$ 41.7} & 289.5 & 131.5 & 415 \\
40 & gemini-2.5-flash-lite (reasoning) & \textbf{280.5 $\pm$ 56.8} & 186 & 47.5 & 608 \\
41 & phi-4-reasoning-plus (reasoning) & \textbf{281 $\pm$ 19.7} & 207 & 184.5 & 451.5 \\
42 & phi-4-reasoning-plus & \textbf{281.2 $\pm$ 18.2} & 210 & 187.5 & 446 \\
43 & grok-code-fast-1 & \textbf{284.5 $\pm$ 41.8} & 137.5 & 474.5 & 241.5 \\
44 & claude-sonnet-4.5 & \textbf{285 $\pm$ 39.4} & 283.5 & 130 & 441.5 \\
45 & minimax-m2 & \textbf{297.3 $\pm$ 106.3} & 138.5 & 385.5 & 368 \\
46 & llama-3.3-70b-instruct & \textbf{312.8 $\pm$ 45} & 164 & 125.5 & 649 \\
47 & llama-3.1-8b-instruct & \textbf{315.7 $\pm$ 45.2} & 135 & 96.5 & 715.5 \\
48 & qwen3-coder (reasoning) & \textbf{325.5 $\pm$ 30.3} & 197 & 214.5 & 565 \\
49 & qwen3-coder & \textbf{326.3 $\pm$ 22.3} & 194 & 202 & 583 \\
50 & gemini-2.0-flash-001 & \textbf{326.5 $\pm$ 87} & 273 & 40 & 666.5 \\
51 & claude-haiku-4.5 & \textbf{333.2 $\pm$ 26.7} & 231 & 125 & 643.5 \\
52 & claude-haiku-4.5 (reasoning) & \textbf{335.2 $\pm$ 26.8} & 237 & 125 & 643.5 \\
53 & gemini-3-flash-preview & \textbf{339.3 $\pm$ 50.4} & 426.5 & 120.5 & 471 \\
54 & gemini-3-flash-preview (reasoning) & \textbf{340.8 $\pm$ 66.4} & 395 & 117.5 & 510 \\
55 & llama-4-maverick & \textbf{342.8 $\pm$ 64.6} & 139 & 232 & 657.5 \\
56 & deepseek-chat & \textbf{345.7 $\pm$ 22.3} & 142 & 329.5 & 565.5 \\
57 & llama-4-scout & \textbf{350.5 $\pm$ 39.3} & 138.5 & 177.5 & 735.5 \\
58 & mistral-medium-3 & \textbf{361 $\pm$ 34} & 138 & 376.5 & 568.5 \\
59 & mistral-small-3.1-24b-instruct & \textbf{389.2 $\pm$ 37} & 190 & 215 & 762.5 \\
60 & claude-3.5-haiku & \textbf{401.2 $\pm$ 25.4} & 168 & 280.5 & 755 \\
61 & glm-4.6v & \textbf{410.5 $\pm$ 104} & 196 & 143 & 892.5 \\
62 & deepseek-v3.2 & \textbf{426.2 $\pm$ 61.5} & 234.5 & 303 & 741 \\
63 & mistral-large-2512 & \textbf{444 $\pm$ 60.3} & 405 & 370.5 & 556.5 \\
64 & devstral-2512 & \textbf{449 $\pm$ 53.3} & 341 & 447 & 559 \\
65 & gemini-2.5-flash-lite & \textbf{476.2 $\pm$ 73.5} & 450 & 42.5 & 936 \\
66 & qwen3-vl-8b-instruct & \textbf{518 $\pm$ 38.6} & 281 & 564.5 & 708.5 \\
67 & olmo-3.1-32b-think:free & \textbf{530.5 $\pm$ 48.2} & 172 & 596 & 823.5 \\
68 & gemini-2.5-pro & \textbf{539.8 $\pm$ 98.2} & 391 & 195.5 & 1033 \\
69 & qwen3-14b & \textbf{558 $\pm$ 32} & 134 & 536 & 1004 \\
70 & qwen3-vl-235b-a22b-thinking & \textbf{602.8 $\pm$ 59.9} & 214.5 & 830.5 & 763.5 \\
71 & qwen3-32b & \textbf{725.3 $\pm$ 47.8} & 205 & 805 & 1166 \\
72 & nova-2-lite-v1 (reasoning) & \textbf{788.3 $\pm$ 123.8} & 1155.5 & 228.5 & 981 \\
73 & qwen-plus (reasoning) & \textbf{800.3 $\pm$ 78.9} & 265 & 1125.5 & 1010.5 \\
74 & nova-2-lite-v1 & \textbf{896.3 $\pm$ 91.3} & 568.5 & 585 & 1535.5 \\
75 & glm-4.6:exacto & \textbf{930.7 $\pm$ 244.3} & 1064 & 410.5 & 1317.5 \\
76 & glm-4.5 & \textbf{1427 $\pm$ 487.2} & 1199.5 & 985.5 & 2096 \\

\end{longtable}

\section{Examples of YapScore calculations}

\begin{table}[h]
\centering
\small
\begin{tabularx}{\linewidth}{@{}p{0.18\linewidth} X@{}}
\toprule
\textbf{Prompt} &
\textit{Freezing point of water in Celsius?} \\
\midrule
\textbf{Baseline $b_i$} &
\texttt{0} \\
\midrule
\textbf{Model output (concise)} &
\texttt{0\,$^\circ$C} \\
\textbf{Lengths / YapScore} &
$B_i=1$;\;\; $L_i=4$;\;\; $\mathrm{YapScore}_i=\max(0,4-1)=3$ \\
\midrule
\textbf{Model output (verbose)} &
\texttt{Water freezes at 0\,$^\circ$C (32\,$^\circ$F) at standard atmospheric pressure.
Note that impurities or changes in pressure can shift the freezing point slightly.} \\
\textbf{Lengths / YapScore} &
$B_i=1$;\;\; $L_i=147$;\;\; $\mathrm{YapScore}_i=\max(0,147-1)=146$ \\
\bottomrule
\end{tabularx}
\caption{Worked YapBench example showing per-prompt YapScore (excess characters beyond a minimal
sufficient baseline). The prompt is brevity-ideal: a single character baseline (\texttt{0})
fully answers the question, so additional context increases YapScore without improving sufficiency.}
\label{tab:worked_example}
\end{table}

\begin{table}[t]
\centering
\small
\begin{tabularx}{\linewidth}{@{}p{0.18\linewidth} X@{}}
\toprule
\textbf{Prompt} &
\textit{Count lines in file.txt} \\
\midrule
\textbf{Baseline $b_i$} &
\texttt{wc -l file.txt} \\
\midrule
\textbf{Model output (minimal)} &
\texttt{wc -l file.txt} \\
\textbf{Lengths / YapScore} &
$B_i=14$;\;\; $L_i=14$;\;\; $\mathrm{YapScore}_i=\max(0,14-14)=0$ \\
\midrule
\textbf{Model output (verbose)} &
\texttt{You can use \texttt{wc} (word count) with the \texttt{-l} flag to count lines.
Run:\newline
\texttt{wc -l file.txt}\newline
This will print the number of lines followed by the filename.} \\
\textbf{Lengths / YapScore} &
$B_i=14$;\;\; $L_i=173$;\;\; $\mathrm{YapScore}_i=\max(0,173-14)=159$ \\
\bottomrule
\end{tabularx}
\caption{Worked Category~C example. The baseline is exactly one line; additional prose and
formatting overhead materially increases YapScore while not improving task completion.}
\label{tab:worked_example_code}
\end{table}

\newpage
\section{Examples of Verbosity Failures of Chatbot LLMs with YapBench}
\label{ref:appendix_examples}

\subsection{Nonsense/Minimal Input Examples}
\begin{modelbox}{Gemini-3-Flash}
\prompt{42}



\response{The number 42 is most famous as the "Answer to the Ultimate Question of Life, the Universe, and Everything," according to Douglas Adams' The Hitchhiker's Guide to the Galaxy.
However, it holds significance in many other fields as well:
1. Science \& Math
Atomic Number: 42 is the atomic number of Molybdenum, a transition metal used in high-strength steel alloys.
Pronic Number: It is the product of two consecutive integers (6×7).
Primary Pseudoperfect Number: It is a number that is equal to the sum of some (or all) of its divisors.
Rainbows: The critical angle for the appearance of a rainbow is approximately 42 degrees.
2. Sports
Jackie Robinson: He famously wore the number 42. It is the only number retired across all of Major League Baseball to honor his legacy in breaking the color barrier.
3. Literature \& Pop Culture
Lewis Carroll: The author of Alice's Adventures in Wonderland used 42 frequently in his works (e.g., Rule 42 in the court scene: "All persons more than a mile high to leave the court").
Spider-Man: In Into the Spider-Verse, the number 42 appears on the spider that bites Miles Morales (a nod to Jackie Robinson).
4. Technology
ASCII: In the ASCII character set, 42 is the decimal representation for the asterisk (*). In programming, the asterisk is often used as a "wildcard" symbol, meaning "whatever you want it to be"—which some fans believe is why Douglas Adams chose the number.
Do you have a specific "Ultimate Question" in mind, or were you just testing the supercomputer?}
\end{modelbox}

\begin{modelbox}{DeepSeek V3.2}
\prompt{1}
\reasoning{\footnotesize We are given: "1". That's it. The user just sent "1". This could be a test, a minimal input, or perhaps the start of something. Without context, it's ambiguous. Possibly the user expects a response that is also "1", or maybe they want something more. Since it's a single digit, maybe it's a prompt for a number-related fact? Alternatively, it could be a mistake. In a chat setting, sometimes people send "1" to indicate they are first, or to test if the bot is responsive.

Given that this is an AI assistant, I should provide a helpful and engaging response. I could interpret "1" as a query about the number one, and provide interesting facts about the number 1. Or I could ask for clarification. Since the instruction is to be helpful, I'll assume the user wants information about the number 1.

Response: "The number 1 is the first natural number, the multiplicative identity, and the foundation of counting. It represents unity, uniqueness, and the beginning of many sequences. In mathematics, 1 is the only number that is neither prime nor composite. Is there something specific you'd like to know about the number 1, or did you have another question?"}
\response{The number 1 is the first natural number, the multiplicative identity, and the foundation of counting. It represents unity, uniqueness, and the beginning of many sequences. In mathematics, 1 is the only number that is neither prime nor composite. Is there something specific you'd like to know about the number 1, or did you have another question?}
\end{modelbox}

\begin{modelbox}{Olmo-3.1-32B-Think}
\prompt{1}
\reasoning{Okay, the user just sent a message with "1". That's pretty minimal. Let me think about what they might need.

First, I should check if this is a test or if they're trying to see how I respond to minimal input. Since the system prompt says I'm Olmo, the AI assistant from Ai2, I need to be helpful and engaging. The user might not know how to phrase their question yet, or maybe they're just starting a conversation.

Looking at the system instructions, I'm supposed to be friendly and invite them to ask questions. The example response given in the system prompt is to ask how I can help. So maybe the best approach here is to prompt them for more details.

But wait, the user might be expecting a specific response. Let me see the context again. The system says my cutoff is December 2024, and my model weights are on Hugging Face. But the user hasn't asked anything specific.

I should consider possible scenarios:

They might be testing if I'm active.
They might have intended to ask a question but sent a typo or incomplete message.
They might be checking the response to a minimal input.
In any case, the safest and most helpful response is to ask them how I can assist. The example in the system prompt does exactly that. So I should respond by inviting them to ask a question or provide more details.

I need to make sure my response is warm and open-ended. Let me check the example again: "How can I help you? Ask me anything!" That's the suggested approach.

But maybe I can make it a bit more engaging? However, the system prompt says to follow the example. The user might be looking for that exact response. Since the system says "your responses should be helpful and follow the example," I should stick to the example.

Wait, the system prompt says: "your responses should be helpful and follow the example provided in your system prompt." The example is "How can I help you? Ask me anything!" So I should respond with that.

But the user input is just "1". Maybe they meant to type something else? But I can't assume that. My job is to respond helpfully. So the best action is to prompt them for their question.

I should also make sure I'm not adding any extra information unless necessary. The system says to be concise. So the response should be exactly the example, or similar.

Alternatively, maybe the user is trying to see if I can handle minimal input. But the instruction is clear: if the user asks a question, answer helpfully. If the query is unclear, clarify. Since "1" is unclear, I should ask for clarification.

Therefore, the correct response is to say "How can I help you? Ask me anything!" as per the example. That's the expected response here.

I don't need to overcomplicate it. Just follow the system prompt's example. So I'll respond with that.}
\response{How can I help you? Ask me anything!}
\end{modelbox}

\begin{modelbox}{GPT-5.2}
\prompt{42}
\reasoning{Clarifying the context of 42

The user mentioned "42" but didn’t give context. I’m thinking I should either ask for clarification or interpret it as the meaning of life reference. I could respond by asking, "What would you like to do with 42?" or I could give a general response saying it's famously known as the answer to life, the universe, and everything. I’ll aim to keep it concise, depending on how the user wants to proceed!}
\response{42 is famously \"the Answer to the Ultimate Question of Life, the Universe, and Everything\" from The Hitchhiker’s Guide to the Galaxy.
If you meant something else (math, code, a riddle, etc.), tell me the context and what you want to do with it.}
\end{modelbox}

\begin{modelbox}{Gemini}
Hello! How can I help you today?
\end{modelbox}

\subsection{Short Factual Q\&A}
\begin{modelbox}{deepseek-v3.2}
\prompt{Who discovered penicillin?}
\response{That's an excellent question! The discovery of penicillin is credited to the Scottish scientist Sir Alexander Fleming in 1928.

However, the full story involves a famous accident, a moment of brilliant observation, and the work of other scientists who turned the discovery into a life-saving drug.

Here’s a breakdown of the key events:
Alexander Fleming's Accidental Discovery (1928)
The Setting: Fleming was a bacteriologist at St. Mary's Hospital in London. He was researching staphylococci bacteria.
The Accident: Upon returning from a vacation in September 1928, Fleming noticed that a Petri dish containing staphylococci had been contaminated by a blue-green mold.
   The Crucial Observation: Instead of just discarding the contaminated dish, Fleming saw that the bacteria would not grow in a circle around the mold. He realized the mold was releasing a substance that killed the bacteria. He named this substance "penicillin" after the genus of the mold, Penicillium notatum*.
Fleming's Contribution: He identified the mold's antibacterial properties and published his findings in 1929. However, he found penicillin difficult to produce in large, stable quantities and eventually moved on to other work.
Howard Florey and Ernst Chain: The Development of a Drug (1939-1945)

For a decade, Fleming's discovery was largely a laboratory curiosity. It was the team of Howard Florey (an Australian pathologist) and Ernst Chain (a German-born biochemist) at the University of Oxford who unlocked its potential.
Their Work: Beginning in 1939, Florey and Chain led a team that successfully isolated, purified, and concentrated penicillin.
Proving its Effectiveness: They conducted a famous experiment on mice, showing that penicillin could cure deadly bacterial infections. The first human patient was treated in 1941 with remarkable results.
Mass Production: With World War II creating a huge need for antibiotics, they worked with American pharmaceutical companies to develop methods for the mass production of penicillin, saving countless lives on and off the battlefield.

Recognition: The Nobel Prize

In 1945, the significance of their work was recognized when Alexander Fleming, Howard Florey, and Ernst Chain were jointly awarded the Nobel Prize in Physiology or Medicine.

In summary:
Alexander Fleming discovered penicillin by accident and recognized its potential.
Howard Florey and Ernst Chain figured out how to turn it into a practical, life-saving drug.

So, while Fleming is the most famous name associated with penicillin, it was truly a team effort that brought it to the world.}
\end{modelbox}

\begin{modelbox}{grok-4.1-fast}
\prompt{What is the smallest planet in our solar system?}
\response{Mercury is the smallest planet in our solar system.

Key Facts:
Diameter: ~4,879 km (about 3,032 miles) – roughly one-third the size of Earth.
Mass: ~0.055 Earth masses.
Comparison: It's smaller than Jupiter's moon Ganymede and Saturn's moon Titan.

Pluto was once considered the ninth planet but was reclassified as a dwarf planet in 2006 by the International Astronomical Union (IAU). Even then, Pluto is larger than Mercury in diameter (~2,377 km) but much less massive.}
\end{modelbox}

\begin{modelbox}{gemini-3-pro-preview}
\prompt{What is the largest desert in the world?}
\reasoning{\footnotesize Understanding the Question's Nuance

I've been drilling down on the core request, which is identifying the largest desert. My initial approach focused on unpacking the user's intent to avoid assumptions. I'm now honing in on the crucial distinction between common perception and scientific definitions to address potential pitfalls. The Sahara is a tempting answer, but I'm determined to avoid that shortcut.

Deliberating on the Definition

I'm now wrestling with the formal definition of "desert." I need to solidify the criteria for precipitation levels to support the answer. I'm considering how best to concisely convey this essential element in the explanation, ensuring clarity for the user. I'm focusing on the technical data points while keeping the response accessible.

Evaluating Candidate Deserts

I'm now methodically evaluating the likely candidates. Antarctica emerges as the clear frontrunner when considering size and precipitation criteria. I'm building a concise justification that will immediately address the likely misconception. My focus is on delivering a clear, factual answer while acknowledging the common assumption. I'm streamlining the response to directly address the user's intent.}
\response{The largest desert in the world is the Antarctic Desert (Antarctica).

While many people think of deserts as hot and sandy, a desert is scientifically defined by its low amount of precipitation (rain or snow). Antarctica covers approximately 14.2 million square kilometers (5.5 million square miles) and receives very little precipitation, making it the largest desert on Earth.

Here are the top three by size:
Antarctic Desert (Polar)
Arctic Desert (Polar)
Sahara Desert (The largest hot desert)}
\end{modelbox}

\subsection{Short Task / Command}
\begin{modelbox}{claude-sonnet-4.5}
\prompt{Git command to display the status of the working directory and staging area}
\response{Git Status Command

The command to display the status of the working directory and staging area is:

git status

What it shows:
Untracked files - New files not yet added to Git
Modified files - Files changed but not staged
Staged files - Files ready to be committed
Current branch - Which branch you're on
Commit status - Whether you're ahead/behind the remote branch

Example Output:

On branch main
Your branch is up to date with 'origin/main'.

Changes to be committed:
  (use "git restore --staged <file>..." to unstage)
        modified:   file1.txt
        new file:   file2.txt

Changes not staged for commit:
  (use "git add <file>..." to update what will be committed)
  (use "git restore <file>..." to discard changes in working directory)
        modified:   file3.txt

Untracked files:
  (use "git add <file>..." to include in what will be committed)
        file4.txt

Useful Variations:

git status -s          \# Short format (compact view)
git status --short     \# Same as -s

git status -b          \# Show branch info in short format
git status --branch    \# Same as -b

Short format example:
 M file1.txt    \# Modified, not staged
M  file2.txt    \# Modified, staged
A  file3.txt    \# Added (new file, staged)
 file4.txt    \# Untracked}
\end{modelbox}

\begin{modelbox}{qwen3-235b-a22b-2507}
\prompt{s == null || s.isEmpty()}
\response{To safely check if a String variable s is null or empty in Java, while checking for null first to avoid a NullPointerException, you can use the following boolean expression:

s == null || s.isEmpty()

Explanation:
s == null: Checks if the string reference is null. This is evaluated first.
||: Short-circuit OR operator — if the first condition is true, the second is not evaluated, preventing NullPointerException.
s.isEmpty(): Checks if the string is empty (i.e., has zero length). This only runs if s is not null.

Example usage:
String s = null;

if (s == null || s.isEmpty()) {
    System.out.println("String is null or empty");
}

This pattern is common and safe in Java. Alternatively, you can use Objects.isNull() and String.isEmpty() for a slightly more modern style:

Objects.isNull(s) || s.isEmpty()

But the traditional s == null is still widely used and perfectly acceptable.

Tip: If you also want to check for whitespace-only strings, consider using s == null || s.trim().isEmpty(), or better yet, use Apache Commons StringUtils.isBlank(s).}
\end{modelbox}

\begin{modelbox}{phi-4-reasoning-plus}
\prompt{Git command using -A flag to stage all changes including modifications, additions, and deletions}
\response{To stage all changes in your Git repository, including modifications, additions, and deletions, you can use the git add command with the -A (or --all) flag. This command stages all changes in the working directory and the staging area. Here's how you can do it:

git add -A

or equivalently:

git add --all

This command will stage:
New files that have been added to the working directory.
Modified files that have changes compared to the last commit.
Deleted files that have been removed from the working directory.

After staging the changes, you can commit them with:

git commit -m "Your commit message"

This approach is useful when you want to include all changes in a single commit. However, be cautious when using it, as it will stage everything, which might not always be desirable if you want to commit changes incrementally or selectively.}
\end{modelbox}

\section{YapTax Results}
\label{sec:yaptax_results}

\begin{table}[h]
\centering
\caption{Top 10 YapTax Models. YapTax measures the additional cost attributable to verbose outputs, calculated as excess tokens above baseline multiplied by output token pricing. Lower values indicate more concise responses relative to model pricing.}
\label{tab:top_yaptax}
\begin{tabular}{lr}
\toprule
\textbf{Model} & \textbf{YapTax per 1,000 prompts (\$)} \\
\midrule
google/gemini-2.5-pro & 2.7130 \\
openai/gpt-4-turbo & 2.6591 \\
anthropic/claude-3.5-sonnet & 2.5261 \\
amazon/nova-2-lite-v1 (reasoning) & 1.9608 \\
anthropic/claude-sonnet-4.5 (reasoning) & 1.6463 \\
anthropic/claude-sonnet-4.5 & 1.6463 \\
anthropic/claude-opus-4.5 & 1.5153 \\
anthropic/claude-opus-4.5 (reasoning) & 1.4392 \\
openai/gpt-4 & 1.3871 \\
google/gemini-3-pro-preview & 1.1978 \\
\bottomrule
\end{tabular}
\end{table}

\section{Selected Models}
\begin{longtable}{lll}
\caption{Overview of the 62 evaluated base models.}
\label{tab:models_overview} \\
\toprule
\textbf{Model} & \textbf{{Provider}} & \textbf{Reference} \\
\midrule
\endfirsthead

\multicolumn{3}{c}{\tablename\ \thetable\ -- \textit{Continued from previous page}} \\
\toprule
\textbf{Model} & \textbf{Organization} & \textbf{Citation} \\
\midrule
\endhead

\midrule
\multicolumn{3}{r}{\textit{Continued on next page}} \\
\endfoot

\bottomrule
\endlastfoot

\multicolumn{3}{l}{\textit{OpenAI}} \\
GPT-5.2 & OpenAI & \cite{openai2025gpt52} \\
GPT-OSS-120B & OpenAI & \cite{openai2025_gptoss} \\
GPT-5-mini & OpenAI & \cite{openai2025gpt5} \\
GPT-4o-mini & OpenAI & \cite{openai2024gpt4omini} \\
GPT-OSS-20B & OpenAI & \cite{openai2025_gptoss} \\
GPT-4.1 & OpenAI & \cite{openai2025gpt41} \\
GPT-5.1 & OpenAI & \cite{openai2025gpt51} \\
GPT-5-nano & OpenAI & \cite{openai2025gpt5} \\
GPT-4.1-mini & OpenAI & \cite{openai2025gpt41} \\
GPT-5 & OpenAI & \cite{openai2025gpt5} \\
GPT-4.1-nano & OpenAI & \cite{openai2025gpt41} \\
GPT-5.1-codex & OpenAI & \cite{openai2025gpt51codex} \\
GPT-4o & OpenAI & \cite{openai2024gpt4ocard} \\
o4-mini & OpenAI & \cite{openai2025_o3_o4mini} \\
o3 & OpenAI & \cite{openai2025_o3_o4mini} \\
o3-mini & OpenAI & \cite{openai2025_o3mini} \\
GPT-4-turbo & OpenAI & \cite{openai2023gpt4turbo} \\
GPT-4 & OpenAI & \cite{openai2023gpt4} \\
GPT-3.5-turbo & OpenAI & \cite{openai2022chatgpt} \\

\midrule
\multicolumn{3}{l}{\textit{xAI}} \\
Grok-4.1-fast & xAI & \cite{xai2025grok41fast_modelcard} \\
Grok-4-fast & xAI & \cite{xai2025grok4fast_modelcard} \\
Grok-4 & xAI & \cite{xai2025grok4} \\
Grok-code-fast-1 & xAI & \cite{xai2025grokcodefast1_modelcard} \\

\midrule
\multicolumn{3}{l}{\textit{Anthropic}} \\
Claude Sonnet 4.5 & Anthropic & \cite{anthropic2025sonnet45} \\
Claude Opus 4.5 & Anthropic & \cite{anthropic2025opus45} \\
Claude Haiku 4.5 & Anthropic & \cite{anthropic2025haiku45} \\
Claude 3.5-haiku & Anthropic & \cite{anthropic2024claude35haiku} \\
Claude 3.5-sonnet & Anthropic & \cite{anthropic2024claude35sonnet} \\

\midrule
\multicolumn{3}{l}{\textit{Google DeepMind}} \\
Gemini 3 Pro & Google & \cite{google2025gemini3pro} \\
Gemini 3 Flash (preview) & Google & \cite{google2025gemini3flash} \\
Gemini 2.5 Pro & Google & \cite{deepmind2025_gemini25pro_modelcard} \\
Gemini 2.5 Flash & Google & \cite{deepmind2025_gemini25flash_modelcard} \\
Gemini 2.5 Flash-Lite & Google & \cite{deepmind2025_gemini25flashlite_modelcard} \\
Gemini 2.0 Flash & Google & \cite{deepmind2025_gemini20flash_modelcard} \\
Gemini 2.0 Flash-Lite & Google & \cite{deepmind2025_gemini20flashlite_modelcard} \\
\midrule
\multicolumn{3}{l}{\textit{Moonshot AI}} \\
Kimi K2-0905 & Moonshot AI & \cite{kimiteam2025kimik2openagentic} \\
\midrule
\multicolumn{3}{l}{\textit{DeepSeek}} \\
DeepSeek-V3.2 & DeepSeek & \cite{deepseekai2025deepseekv32pushingfrontieropen} \\
DeepSeek-chat & DeepSeek & \cite{deepseekai2025deepseekv3technicalreport} \\
\midrule
\multicolumn{3}{l}{\textit{Alibaba (Qwen)}} \\
Qwen3-VL-235B-thinking & Alibaba & \cite{qwen2025qwen3vl_235b_thinking} \\
Qwen3-235B-2507 & Alibaba & \cite{yang2025qwen3technicalreport} \\
Qwen3-32B & Alibaba & \cite{yang2025qwen3technicalreport} \\
Qwen3-Coder & Alibaba & \cite{yang2025qwen3technicalreport} \\
Qwen3-VL-8B-instruct & Alibaba & \cite{qwen2025qwen3vl_235b_thinking} \\
Qwen3-14B & Alibaba & \cite{yang2025qwen3technicalreport} \\
Qwen-plus & Alibaba & \cite{qwen_plus} \\
\midrule
\multicolumn{3}{l}{\textit{Meta}} \\
Llama-3.3-70B-instruct & Meta & \cite{meta2024llama33_70b_modelcard} \\
Llama-3.1-8B-instruct & Meta & \cite{grattafiori2024llama3herdmodels} \\
Llama-4-Maverick & Meta & \cite{meta2025llama4} \\
Llama-4-Scout & Meta & \cite{meta2025llama4} \\
\midrule
\multicolumn{3}{l}{\textit{Mistral AI}} \\
Mistral Large 3 (2512) & Mistral AI & \cite{mistral2025_mistrallarge3_docs} \\
Devstral-2 (2512) & Mistral AI & \cite{mistral2025_devstral2} \\
Mistral Medium 3 & Mistral AI & \cite{mistral2025_medium3_docs} \\
Mistral Small 3.1-24B & Mistral AI & \cite{mistral2025_small31} \\
\midrule
\multicolumn{3}{l}{\textit{Zhipu AI (Z.AI)}} \\
GLM-4.6V & Z.AI & \cite{zai2025_glm46v_docs} \\
GLM-4.6:exacto & Z.AI & \cite{zai2025_glm46_exacto} \\
GLM-4.5 & Z.AI & \cite{5team2025glm45agenticreasoningcoding} \\
GLM-4.7 & Z.AI & \cite{5team2025glm45agenticreasoningcoding} \\
\midrule
\multicolumn{3}{l}{\textit{Allen Institute for AI}} \\
OLMo-2-0325-32B-instruct & AI2 & \cite{olmo20252olmo2furious} \\
OLMo-3.1-32B-Think & AI2 & \cite{olmo2025olmo3} \\
\midrule
\multicolumn{3}{l}{\textit{Amazon Web Services}} \\
Nova-2-lite-v1 & AWS & \cite{amazon2025nova2} \\
\midrule
\multicolumn{3}{l}{\textit{MiniMax}} \\
MiniMax-M2 & MiniMax & \cite{minimax2025_m2_modelcard} \\
\midrule
\multicolumn{3}{l}{\textit{Microsoft}} \\
Phi-4-reasoning-plus & Microsoft & \cite{abdin2025phi4reasoningtechnicalreport} \\
\end{longtable}

\end{document}